%% file: icra20.tex
\documentclass[conference]{IEEEtran}
\IEEEoverridecommandlockouts
    
\usepackage{bm}
\usepackage{url}
\usepackage{tikz}
\usepackage{soul}
\usepackage{cite}
\usepackage{color}
\usepackage{pifont}
\usepackage{caption}
\usepackage{booktabs}
\usepackage{textcomp}
\usepackage{setspace}
\usepackage{footnote}
\usepackage{verbatim}
\usepackage{microtype}
\usepackage{algorithm}
\usepackage{graphicx}
\usepackage{epstopdf}
\usepackage{multirow}
\usepackage{multicol}
\usepackage{algorithmicx}
\usepackage[most]{tcolorbox}
\usepackage{algpseudocode}
\algrenewcommand\textproc{} 
\usepackage[export]{adjustbox}
\usepackage{amssymb, amsmath, times}
\usepackage[left=.78in,top=1in,right=.78in,bottom=.9in,headsep=0.25in,headheight=20pt,heightrounded]{geometry}
\usepackage[skip=0cm,list=true,labelfont=it]{subcaption}
\usepackage[pagebackref=true,breaklinks=true,colorlinks=true, citecolor=green,linkcolor=cyan,urlcolor=blue,bookmarks=true]{hyperref}
\def\BibTeX{{\rm B\kern-.05em{\sc i\kern-.025em b}\kern-.08em
		T\kern-.1667em\lower.7ex\hbox{E}\kern-.125emX}}

\DeclareGraphicsExtensions{.pdf,.jpeg,.png,.eps}

\begin{document}
\bstctlcite{IEEEexample:BSTcontrol}
\title{Mechanism and Model of a Soft Robot for Head Stabilization in Cancer Radiation Therapy.}
\author{Olalekan Ogunmolu$^{\dagger}$, Xinmin Liu$^{\star}$, Nicholas Gans$^{\ddagger}$, and  Rodney D. Wiersma$^{\dagger}$
	\thanks{$^{\dagger}$Perelman School of Medicine,
		The University of Pennsylvania, Philadelphia, PA 19104, USA.
		{\tt\small \{olalekan.ogunmolu, rodney.wiersma\}@pennmedicine.upenn.edu}}%
	\thanks{$^{\star}$Department of Radiation and Cellular Oncology, The University of Chicago, Chicago, IL 60637, USA.
		{\tt\small xmliu@uchicago.edu}
	}%
	\thanks{$^{\ddagger}$Automation \& Intelligent Systems Division, The University of Texas at Arlington Research Institute, Arlington, TX, USA.
		{\tt\small nick.gans@uta.edu}
}%
	\thanks{The research reported in this publication was supported by National Cancer Institute of the National Institutes of Health under award number R01CA227124.}
}

\maketitle

\input{sections/def}
\input{sections/abstract}
\input{sections/intro}
\input{sections/mech}

\input{sections/analysis}
\input{sections/simulation}
\input{sections/conclusion}
\input{sections/acks}
\bibliographystyle{IEEEtran}
\bibliography{biblio}	 	
\end{document}

%% file: sections/def.tex
\definecolor{light-blue}{rgb}{0.30,0.35,1}
\definecolor{light-green}{rgb}{0.20,0.49,.85}
\definecolor{purple}{rgb}{0.70,0.69,.2}

\newcommand{\lb}[1]{\textcolor{light-blue}{#1}}
\newcommand{\bl}[1]{\textcolor{blue}{#1}}

\newcommand{\maybe}[1]{\textcolor{gray}{\textbf{MAYBE: }{#1}}}
\newcommand{\inspect}[1]{\textcolor{cyan}{\textbf{CHECK THIS: }{#1}}}
\newcommand{\more}[1]{\textcolor{red}{\textbf{MORE: }{#1}}}
\renewcommand{\figureautorefname}{Figure}
\renewcommand{\sectionautorefname}{$\S$}
\renewcommand{\equationautorefname}{equation}
\renewcommand{\subsectionautorefname}{$\S$}
\renewcommand{\chapterautorefname}{Chapter}

\newcommand{\cmt}[1]{{\footnotesize\textcolor{red}{#1}}}
\newcommand{\todo}[1]{\textcolor{cyan}{TO-DO: #1}}
\newcommand{\lekan}[1]{\cmt{\textbf{LO}: #1}}
\newcommand{\nick}[1]{\cmt{\textbf{NG}: #1}}
\newcommand{\mike}[1]{\cmt{\textbf{Mike}: #1}}
\newcommand{\steve}[1]{\cmt{\textbf{SJ}: #1}}
\newcommand{\review}[1]{\noindent\textcolor{red}{$\rightarrow$ #1}}
\newcommand{\response}[1]{\noindent{#1}}
\newcommand{\stopped}[1]{\color{red}STOPPED HERE #1\hrulefill}

\newcounter{mnote}
\newcommand{\marginote}[1]{\addtocounter{mnote}{1}\marginpar{\themnote. \scriptsize #1}}
\setcounter{mnote}{0}
\newcommand{\ie}{$i.e.$\ }
\newcommand{\eg}{e.g.\ }
\newcommand{\cf}{c.f.\ }
\newcommand{\yes}{\checkmark}
\newcommand{\no}{\ding{55}}

\newcommand{\flabel}[1]{\label{fig:#1}}
\newcommand{\seclabel}[1]{\label{sec:#1}}
\newcommand{\tlabel}[1]{\label{tab:#1}}
\newcommand{\elabel}[1]{\label{eq:#1}}
\newcommand{\alabel}[1]{\label{alg:#1}}
\newcommand{\fref}[1]{\cref{fig:#1}}
\newcommand{\sref}[1]{\cref{sec:#1}}
\newcommand{\tref}[1]{\cref{tab:#1}}
\newcommand{\eref}[1]{\cref{eq:#1}}
\newcommand{\aref}[1]{\cref{alg:#1}}

\newcommand*\idx[2][]
{
\def\next{#1}%
\ifx\empty\next
  (#2)
\else
  (#1, #2)
\fi
}
\newcommand*\elt[3][]
{
\def\next{#1}%
\ifx\empty\next
  #2\idx{#3}
\else
  #1\idx{#2,#3}
\fi
}
\newcommand*\pd[3][]
{
\def\next{#1}%
\ifx\empty\next
  \frac{\partial#2}{\partial #3}
\else
  \frac{{\partial^{#1} #2}}{\partial#3^{#1}}
\fi
}
\newcommand*\pdn[3]
{
\frac{{\partial#1}^{#3}}{\partial^{#3} #2}
}
\newcommand{\bull}[1]{$\bullet$ #1}
\newcommand{\argmax}{\text{argmax}}
\newcommand{\argmin}{\text{argmin}}
\newcommand{\mc}[1]{\mathcal{#1}}
\newcommand{\bb}[1]{\mathbb{#1}}
\newcommand{\deq}{\mathrel{\stackrel{\text{\tiny{def}}}{=}}}
\newcommand{\opequals}[1]{\ #1\hspace{-3pt}=\hspace{1pt}}
\newcommand{\plusequals}{\opequals{+}}
\newcommand{\minusequals}{\opequals{-}}
\newcommand{\timesequals}{\opequals{*}}
\newcommand{\front}[1]{beg(#1)}
\newcommand{\back}[1]{end(#1)}
\newcommand{\stddev}{\sigma}
\newcommand{\variance}{\stddev^2}
\newcommand{\ifelse}[3]{\begin{cases}#1 \text{ if } #2\\#3 \text{ otherwise}\end{cases}}

\newcommand{\figdir}{figures/}
\newcommand{\capt}[2]{\caption[#1]{#1#2}}
\newcommand{\qcapt}[1]{\capt{#1}{}}
\newcommand{\figheadingnospace}[1]{\center{\textbf{#1}}}
\newcommand{\figheading}[1]{\figheadingnospace{#1}\vspace{3mm}}
\newcommand{\fig}[5]
{
\begin{figure}
\begin{center}
\includegraphics[width=#3\columnwidth]{figures/#1}
\end{center}
\capt{#4}{#5}
\flabel{#2}
\end{figure}
}
\newcommand{\figtbph}[5]
{
\begin{figure}[tbph]
\begin{center}
\includegraphics[width=#3\columnwidth]{figures/#1}
\end{center}
\capt{#4}{#5}
\flabel{#2}
\end{figure}
}
\newcommand{\figt}[5]
{
\begin{figure}[tb]
\begin{center}
\includegraphics[width=#3\columnwidth]{figures/#1}
\end{center}
\capt{#4}{#5}
\label{#2}
\end{figure}
}
\newcommand{\figstar}[5]
{
\begin{figure*}
\begin{center}
\includegraphics[width=#3\textwidth]{figures/#1}
\end{center}
\capt{#4}{#5}
\flabel{#2}
\end{figure*}
}
\newcommand{\qfig}[3]{\fig{#1}{#1}{#2}{#3}{}}

\def\tidx{t}
\newcommand{\Note}[1]{}
\renewcommand{\Note}[1]{\hl{[#1]}}  
\newcommand{\NoteSigned}[3]{{\sethlcolor{#2}\Note{#1: #3}}}
\newcommand{\NoteLO}[1]{\NoteSigned{LO}{YellowGreen}{#1}}  
\newcommand{\NoteDN}[1]{\NoteSigned{DN}{LightBlue}{#1}}    


\newcommand{\bolditalic}[1]{\textbf{#1}}

\def\control{\textbf{u}}
\def\action{\text{\textit{u}}} 
\def\state{\text{\textbf{y}}} 
\def\netstate{s}
\def\statenext{\text{\textit{s}}_{\tidx+1}} 
\def\reward{\text{\textit{r}}} 
\def\dosegrid{4^\circ}
\def\actionspace{\mc{U}}
\def\buf{\mc{W}}
\def\capmax{\mc{M}}
\def\patient{\text{\textit{p}}}
\def\patients{\text{\textit{P}}}
\def\slice{\text{\textit{N}}}
\def\beamangle{\theta}
\def\beamlets{\mc{B}}
\def\dosemat{\mc{D}}
\def\voxtot{\mc{X}}
\def\dij{\dosemat_{ij}}
\def\beamsall{\bm{\Theta}}
\def\experience{\bf{\textit{e}}}
\def\qfunc{\textit{Q}}
\def\neti{\bm{f_\psi}}
\def\netii{g_\phi}
\def\bixel{x}
\def\acti{\action_\tidx}
\def\actii{\action_\tidx^2}
\def\optimalval{v^\star(\textit{s})}
\def\beamblock{\textbf{\textit{B}}}
\def\featureplanes{\textbf{\textit{X}}}
\def\dosemask{\textbf{\textit{D}}}
\def\lastfeatureplane{\textbf{\textit{E}}}
\def\featurebinmask{\textbf{\textit{Y}}}

\def\stateset{\mc{X}}
\def\cost{Q} 
\def\basis{\bm{e}}
\def\policy{\bm{\pi}}
\def\policyimproved{\bar{\policy}}
\def\policyspace{\bm{\Pi}}
\def\visitcount{\mathds{N}(\netstate, \action)}
\def\netprobs{P(\state,	\action)}

\def\valuenet{v_\psi(\state)}
\def\probneti{p(\netstate, \action)}

\def\treepol{\bm \pi_z}
\def\explorefactor{c}
\def\uval{U(\netstate, \action)}
\def\node{\textbf{\textit{x}}}
\def\newnode{\bar \node}
\def\pointerparent{\textit{\node.p}}
\def\pointerchild{\textit{\node.child}}

\def\lagrangian{\mc{L}}
\def\primal{\textbf{\textit{x}}} 
\def\admmvar{\textbf{\textit{z}}}
\def\dual{\bm{\lambda}}
\def\Amat{\textbf{\textit{A}}}
\def\Bmat{\textbf{\textit{B}}}
\def\cvec{\textbf{\textit{c}}}
\def\admmpen{\rho}
\def\kldiv{\textbf{D}_{KL}}

\def\game{\Gamma}

\newcommand{\subbold}[1]
{
	_\textbf{#1}
}

\newcommand{\supbold}[1]
{
	^\textbf{#1}
}

\newcommand{\xbold}[1]
{
	\textbf{#1}
}

\newcommand{\subsupbold}[3]
{
	\textbf{#1} _{#2}^{#3}
}
\newcommand{\hatsubsupbold}[3]
{
	\hat{\textbf{#1}} _{#2}^{#3}
}

\newcommand{\symb}[1]
{
	${\textbf{#1}}$
}

\newcommand{\sym}[1]{$#1$}
\newcommand{\eq}[2]
{
	\begin{equation}  \label{#2}
	\begin{split}  
	#1
	\end{split}
	\end{equation}
}
\newcommand{\gam}{\bm{\Gamma}}
\newcommand{\lamb}{\bm{\Lambda}}
\newcommand{\neterror}{\bm{\varepsilon}_f}
\newcommand{\lyap}{\textbf{V}}
\newcommand{\lyapder}{\dot{\textbf{V}}}
\newcommand{\processnoise}{\textbf{w}(k)}
\newcommand{\stateT}{\textbf{y}^T}
\newcommand{\gainx}{\subsupbold{K}{y}{}}
\newcommand{\gainxT}{\subsupbold{K}{y}{^T}}
\newcommand{\gainxdiff}{\tilde{\xbold{K}}_y^T}
\newcommand{\gainxdiffo}{\tilde{\xbold{K}}_y}
\newcommand{\gainrdiff}{\tilde{\xbold{K}}_r^T}
\newcommand{\gainrdiffo}{\tilde{\xbold{K}}_r}
\newcommand{\gainr}{\subsupbold{K}{r}{}}
\newcommand{\gainrT}{\subsupbold{K}{r}{^T}}
\newcommand{\gainxest}{\dot{\hat{\textbf{K}}}_y}
\newcommand{\gainrest}{\dot{\hat{\textbf{K}}}_r}
\newcommand{\errorest}{\dot{\hat{\epsilon}}_f}
\newcommand{\gainxesto}{\hat{\textbf{K}}_y}
\newcommand{\gainresto}{\hat{\textbf{K}}_r}
\newcommand{\gammax}{\bm{\Gamma_y}}
\newcommand{\gammaxT}{\bm{\Gamma_y}^T}
\newcommand{\gammar}{\bm{\Gamma}_r}
\newcommand{\gammarT}{\bm{\Gamma}_r^T}

\newcommand{\ttbf}[1]{\textbf{\texttt{#1}}}
\newcommand{\ttit}[1]{\texttt{#1}}

\newcommand{\domain}[1]{\mathcal{#1}}

\def\nomcontrol{\bar{\textbf{u}}}
\def\nomdist{\bar{\textbf{v}}}
\def\control{\textbf{u}}
\def\admcontrol{\mathcal{U}}
\def\nomstate{\bar{\textbf{x}}}
\def\state{\textbf{x}}
\def\admstate{\mathcal{x}}
\def\stateT{\textbf{x}^T}
\def\policy{\bm{\pi}}
\def\param{\bm{\theta}}
\def\reward{r}
\def\sumreward{\domain{R}}
\def\augreward{\ell}
\def\augsumreward{\ell}
\def\traj{\bm{\tau}}
\def\trajboth{\bar{\bm{\tau}}}
\def\hinf{H_\infty}
\def\polstoch{\bm{\pi_\param}(\control_t| \state_t)}
\def\polstochadv{\bm{\pi_\param}(\distbo_t| \state_t)}
\def\polstochaug{\bm{\pi_\param}(\control_t |\state_t)}
\def\polstochaugall{\bm{\pi}_{\param i}(\control_t | \state_t)}
\def\sensy{\bm{\gamma}}
\def\dist{\textbf{v}}
\def\distbo{\textbf{v}}
\def\noise{\textbf{w}}
\def\entropy{\domain{H}}
\def\controlcombo{p_{uvi}}
\def\protcontrol{p(\control_k | \state_k)}
\def\protcontrolall{p_i(\control_k | \state_k)}
\def\advcontrol{p(\distbo_k | \state_k)}
\def\advcontrolall{p_i(\distbo_k | \state_k)}
\def\jointlocalpolicy{p(\control_k | \distbo_k, \state_k)}
\def\jointlocalpolicyall{p_i(\control_k | \distbo_k, \state_k)}


\def\hinf{H_\infty}
\def\tidx{t}
\def\fstate{f_{\state \tidx}}
\def\fcontrol{f_{\control \tidx}}
\def\fcontrolcontrol{f_{\control\control \tidx}}
\def\fcontroldistbo{f_{\control\distbo \tidx}}
\def\fdistbo{f_{\distbo \tidx}}
\def\fdistbodistbo{f_{\distbo \distbo \tidx}}
\def\fdistbostate{f_{\distbo \state \tidx}}
\def\fcontrolstate{f_{\control \state \tidx}}
\def\Vstate{V_{\state \tidx+1}}
\def\Vstateback{V_{\state \tidx}}
\def\Vstatestate{V_{\state \state \tidx+1}}
\def\Vstatestateback{V_{\state \state \tidx}}
\def\Qstate{Q_{\state \tidx}}
\def\Qcontrol{Q_{\control \tidx}}
\def\Qdistbo{Q_{\distbo \tidx}}
\def\Qstatestate{Q_{\state \state \tidx}}
\def\Qstatecontrol{Q_{\state \control \tidx}}
\def\Qcontrolstate{Q_{\control \state \tidx}}
\def\Qcontrolcontrol{Q_{\control \control \tidx}}
\def\Qstatecontrol{Q_{\state \control \tidx}}
\def\Qcontrolstate{Q_{\control \state \tidx}}
\def\Qcontrolcontrol{Q_{\control \control \tidx}}
\def\Qcontroldistbo{Q_{\control \distbo \tidx}}
\def\Qdistbostate{Q_{\distbo \state \tidx}}
\def\Qdistbodistbo{Q_{\distbo \distbo \tidx}}
\def\Qstatedistbo{Q_{\state \distbo \tidx}}
\def\Qdistbostate{Q_{\distbo \state \tidx}}
\def\Qdistbocontrol{Q_{\distbo \control \tidx}}
\def\TQstate{\tilde{Q}_{\state \tidx}}
\def\TQcontrol{\tilde{Q}_{\control \tidx}}
\def\TQdistbo{\tilde{Q}_{\distbo \tidx}}
\def\TQstatestate{\tilde{Q}_{\state \state \tidx}}
\def\TQstatecontrol{\tilde{Q}_{\state \control \tidx}}
\def\TQcontrolstate{\tilde{Q}_{\control \state \tidx}}
\def\TQcontrolcontrol{\tilde{Q}_{\control \control \tidx}}
\def\TQstatecontrol{\tilde{Q}_{\state \control \tidx}}
\def\TQcontrolstate{\tilde{Q}_{\control \state \tidx}}
\def\TQcontrolcontrol{\tilde{Q}_{\control \control \tidx}}
\def\TQcontroldistbo{\tilde{Q}_{\control \distbo \tidx}}
\def\TQdistbostate{\tilde{Q}_{\distbo \state \tidx}}
\def\TQdistbodistbo{\tilde{Q}_{\distbo \distbo \tidx}}
\def\TQstatedistbo{\tilde{Q}_{\state \distbo \tidx}}
\def\TQdistbostate{\tilde{Q}_{\distbo \state \tidx}}
\def\TQdistbocontrol{\tilde{Q}_{\distbo \control \tidx}}

\def\cost{\mathcal{J}} 
\def\lstate{\ell_{\state \tidx}}
\def\lcontrol{\ell_{\control \tidx}}
\def\ldistbo{\ell_{\distbo \tidx}}
\def\lstatestate{\ell_{\state \state \tidx}}
\def\lstatecontrol{\ell_{\state \control \tidx}}
\def\lcontrolstate{\ell_{\control \state \tidx}}
\def\lcontrolcontrol{\ell_{\control \control \tidx}}
\def\lstatecontrol{\ell_{\state \control \tidx}}
\def\lcontrolstate{\ell_{\control \state \tidx}}
\def\lcontrolcontrol{\ell_{\control \control \tidx}}
\def\lcontroldistbo{\ell_{\control \distbo \tidx}}
\def\ldistbostate{\ell_{\distbo \state \tidx}}
\def\ldistbodistbo{\ell_{\distbo \distbo \tidx}}
\def\lstatedistbo{\ell_{\state \distbo \tidx}}
\def\ldistbostate{\ell_{\distbo \state \tidx}}
\def\ldistbocontrol{\ell_{\distbo \control \tidx}}

\def\delstate{\delta \state}
\def\delcontrol{\delta \control}
\def\delnoise{\delta \noise}
\def\deldistbo{\delta \distbo}

\def\Kbutt{\Qcontrolcontrol^T \Qdistbodistbo - \Qcontroldistbo^T \Qcontroldistbo}

\def\vectorgain{\textbf{\textit{g}}}
\def\matrixgain{\textbf{\textit{G}}}

\def\optimcont{\pi_{\control}^\star}
\def\optimdistbo{\pi_{\distbo}^\star}
\def\udist{p_{\control_i}}
\def\vdist{p_{\distbo_i}}
\def\param{\bm{\theta}}
\def\blim{B_{\text{lim}}}
\def\ball{B_{\text{tot}}}


\def\coriolis{\textbf{\textit{C}}}
\def\massinertia{\textbf{\textit{M}}}
\def\torque{\bm{\tau}}
\def\frictionvec{\textbf{\textit{f}}}
\def\Smat{\textbf{\textit{S}}}
\def\Bmat{\textbf{\textit{B}}}
\def\wheelrad{\textbf{\textit{r}}}

\def\stateweight{\textbf{\textit{w}}_x}
\def\actionweight{\textbf{\textit{w}}_u}
\def\advactionweight{\textbf{\textit{w}}_v}

\def\kau{\mc{K}}
\def\particle{\textbf{x}}
\def\deformationgrad{\textbf{F}}
\def\refconf{\bm{\bm{\chi}}_0}
\def\refconfbody{\mathscr{B}_0}
\def\conf{\bm{\bm{\chi}}}
\def\currconf{\bm{\chi}}
\def\Eulerian{\mc{E}}
\def\cauchystress{\bm{\sigma}}
\def\stresscomp{\sigma}
\def\currconfbody{\mathscr{B}}
\def\strain{W}
\def\materialresponse{\textbf{G}}
\def\orthoggroup{{\textit{SO}}(3)}
\def\liegroup{{\textit{SE}}(3)}
\def\liealgebra{\mathfrak{se}(3)}
\def\identity{\textbf{I}}
\newcommand{\trace}[1]{\textbf{tr}(#1)}
\def\leftcauchy{\textbf{B}}
\def\rightcauchy{\textbf{C}}
\def\fiber{d\textbf{X}}

\def\dof{\text{DOF }}
\def\dofs{\text{DOFs }}
\def\reline{\mathbb{R}}
\def\curve{\deformationgrad}
\def\twist{{\xi}}
\def\contactforce{\tilde{F}}
\def\contactforcecomp{f}
\def\gaussianmap{\textbf{}n}
\def\contacttorquecomp{\tau}
\def\wrt{with respect to }
\def\curveparam{\position}
\def\pose{{g}}
\def\selmap{{B}}
\def\manipmap{{G}}
\def\jacob{\bm{J}}
\def\position{\textbf{r}}
\def\deformationgradcur{\textbf{H}}
\def\eulerianvel{\textbf{v}(\position, t)}
\def\headparam{\zeta}

\newcommand{\putsoro}[2]{\includegraphics[width=.45\columnwidth,height=#2\columnwidth]{../../../PhDThesis/figures/#1}}
\newcommand{\sorowidth}{.35}

\def\qfunc{\textit{Q}}
\def\game{\Gamma}
\def\basepolicy{\{\mu_0, \ldots, \mu_{N-1}\}}

%% file: sections/abstract.tex
\begin{abstract}
We present a parallel robot mechanism and the constitutive  laws that govern the deformation of its constituent soft actuators. Our ultimate goal is the real-time motion-correction of a patient's head  deviation from a target pose where the soft actuators control the position of the patient's cranial region on a treatment machine. 
We describe the mechanism, derive the stress-strain constitutive laws for the individual actuators and the inverse kinematics that prescribes a given deformation, and then present simulation results that validate our mathematical formulation. Our results  demonstrate deformations consistent with our radially symmetric displacement formulation under a finite elastic deformation framework. 
\end{abstract}

%% file: sections/intro.tex
\section{INTRODUCTION}
Along with chemotherapy and surgery, radiation therapy (RT) is an effective method of cancer treatment,  with more than half of all cancer patients managed by RT having higher survival rates~\cite{Gaspar}. This is in part due to the technological advancements that enable maximizing radiation dose to a tumor target, whilst simultaneously minimizing radiation to surrounding healthy tissues within a target volume. 

\begin{figure}[tb!]
	\centering
	\begin{tabular}{@{}c@{}c@{}}
		\includegraphics[width=0.24\textwidth]{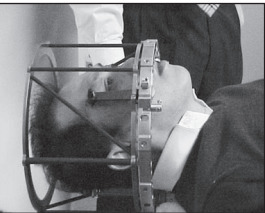} & 
		\hfill 
		\includegraphics[width=0.240\textwidth, height=.21\textwidth]{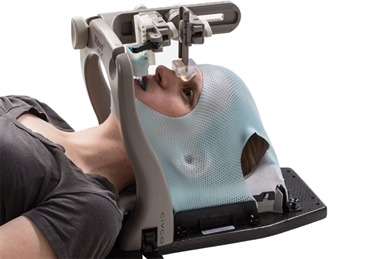} 
	\end{tabular}
	\caption{\textit{Left:} The Brown-Robert-Wells SRS Head Frame, reprinted from \cite{CranialBetter}. \textit{Right:} Thermoplastic face mask.}
	\label{fig:rigid_cyber}
\end{figure}

To assure optimal dose delivery, the patient lies supine on a treatment table and their position with respect to the treatment machine must not exceed submillimeter and subdegree deviations from treatment targets. The current clinical convention for RT and stereotactic radiosurgery (SRS) is to immobilize the patient with rigid metallic frames or masks (see \autoref{fig:rigid_cyber}). However, frames attenuate the radiation dose (thus lowering treatment quality) owing to their metallic components, lack real-time motion compensation (hence the need for stopping the treatment when the patient deviates from a target position beyond a given threshold), and they cause patient discomfort and pain owing to their invasiveness~\cite{Takakura}. The limitations of frames have spurred clinics to use thermoplastic face masks. The masks decrease accuracy because of their  flex, which can cause a drift of up to 2-6mm; shrink and deformation in the mask's physical structure are also prominent, arising from repeated use.  
The motion correction precisions provided by frames and masks are not suitable for deep tumors located near critical structures such as the brain stem or for newer treatment modalities such as single isocenter multiple-target stereotactic radiosurgery (SRS), which are highly sensitive to rotational head motions. 


%

To overcome these issues, explorative robotic positioning research studies have demonstrated the feasibility of maintaining stable patient cranial motion consistent with treatment plans~\cite{BelcherThesis, Xinmin4DOF, XinminPlosOne, HerrmannHexaPODMPC}. For example, Belcher et al's Stewart-Gough platform~\cite{BelcherThesis}   achieves $\le 0.5$mm and $\le 0.5^\circ$ positioning accuracy $99\%$ of the time. This system, along with the plastic-based Ostyn et al's Stewart-Gough platform~\cite{Ostyn17}, use stepper motors to actuate the robot links. 
The 6-\dof robotic HexaPOD treatment couch of~\cite{HerrmannHexaPODMPC} was used in lung tumors treatment evaluation. Leveraging the fast and precise positioning of heavy payloads, the authors implemented a linear auto-regressive exogenous parameter-identification system to identify the HexaPOD's dynamics. The authors of~\cite{CouchModelingSimulation} used an Elekta 4-\dof (3 translation and one rotational) parallel robot to first simulate and then control couch-based motion in real-time. A linear state-space model approximated the rigid body dynamics of the patient support system they had proposed in~\cite{Haas2005}. 
These  systems are accompanied by the following hazards:
\begin{itemize}
	\item they share their dextrous workspace with the patients' body -- a safety concern since these robots' rigid mechanical components are non-compliant;  
	\item their lack of structural compliance mean that the patient experiences ``hard shocks" when the end effector moves; 
	\item they are incapable of providing sophisticated motion compensation that may  be needed for respiratory and internal organs displacement that often cause deviation from the target; and
	\item their component electric motors and linear actuators introduce radiation attenuation and serious safety concerns.  
\end{itemize}  
%

%
\subsection{Contributions}
This is why we have proposed inflatable air bladders (IABs) as motion compensators during treatment (see ~\cite{Gu15AAPM,Ogunmolu15CASE,Ogunmolu16CASE,Ogunmolu17IROS, OgunmoluThesis, Ogunmolu19Springer}). Contrary to stochastic system identification methods used in deriving our earlier models, here, we regulate volume fractions within the IABs and their spatial deformation based on specific mathematical relationships. Furthermore, we carve out a new class of IABs that are continuum, compliant, and configurable (C3) soft actuators that provide therapeutic patient head motion compensation. Contrary to remote-controlled airbags that have been used in upper mandible and head manipulation~\cite{ishizaka2014remote}, our actuators deform based on their material moduli, compressed air pressurization and incompressibility constraints when given a reference trajectory. 
%
%
%
Specifically, our \textbf{contributions} are as follows:

\begin{itemize}
	\item  We propose a minimally-invasive mechanism that largely avoids dose attenuation, whilst providing patient comfort in real-time head motion correction;
	\item We derive a constitutive model for the robot's constituent actuators by extending the principles of nonlinear elastic deformations~\cite{OgdenBook, Treloar1975} to strain deformations;
	\item We analyze their deformation under stress, strain, internal pressurization, and an arbitrary hydrostatic pressure. 
\end{itemize}
The rest of this paper is structured as follows: in \autoref{sec:mechanism}, we present the overall C3 kinematic mechanism; we analyze the deformation properties of the IAB in \autoref{sec:deform}; we then provide and discuss simulation results in \autoref{sec:simu}. We conclude the paper in \autoref{sec:conclude}.

%% file: sections/mech.tex
\section{Mechanism Synthesis}
\label{sec:mechanism}

In our previous IAB models~\cite{Ogunmolu17IROS}, we used a system identification approach to realize the overall system model. Our resultant model lumped the patient, treatment couch, as well as IAB models into one. The disadvantage of this approach was that such overall model lacked sufficient fidelity such that it necessitated the  memory-based adaptive control composite laws that were derived from inverse Lyapunov analysis. Furthermore, the approximation model component of the ensuing neural-network controller required extensive training to realize a suitable controller for our head immobilization.  Our goal here is to realize closed-form constitutive models for the IABs -- capable of manipulating the complete patient's head  \dof motion in real-time. 

\subsection{Type Synthesis}
To design a soft mechanism that precisely manipulates a patient on a treatment table, optimizing the geometrical synthesis leaves many imponderables  unresolved  given the multiple choices that arise, each calling for careful judgment in weighing advantages against disadvantages. Owing to the success of parallel mechanisms in precision manipulation tasks~\cite{Merlet2015}, we decide upon a profile-mechanism consisting of spherically-symmetric soft actuators arranged in a parallel manner around the patient's skull. We recognize that other kinematicians may arrive at other linkage mechanisms that may offer better results. Our goal here is to find a high-fidelity model with tractable kinematics that can move the patient's head as desired on  a treatment table. 

\subsection{Number Synthesis} 
\label{subsec:num_syn}
%
We seek such freedoms and constraints in the structural properties of the mechanism that enables rapid motion correction when a patient deviates from target. Prehensile control of the patient's cranial motion is attractive given its erstwhile success, \eg ~\cite{ishizaka2014remote}'s airbag mechanism. 
\begin{figure}[!tb]
	\centering
	\begin{minipage}[b]{.24\textwidth}
		\includegraphics[width=\textwidth,height=.9\textwidth]{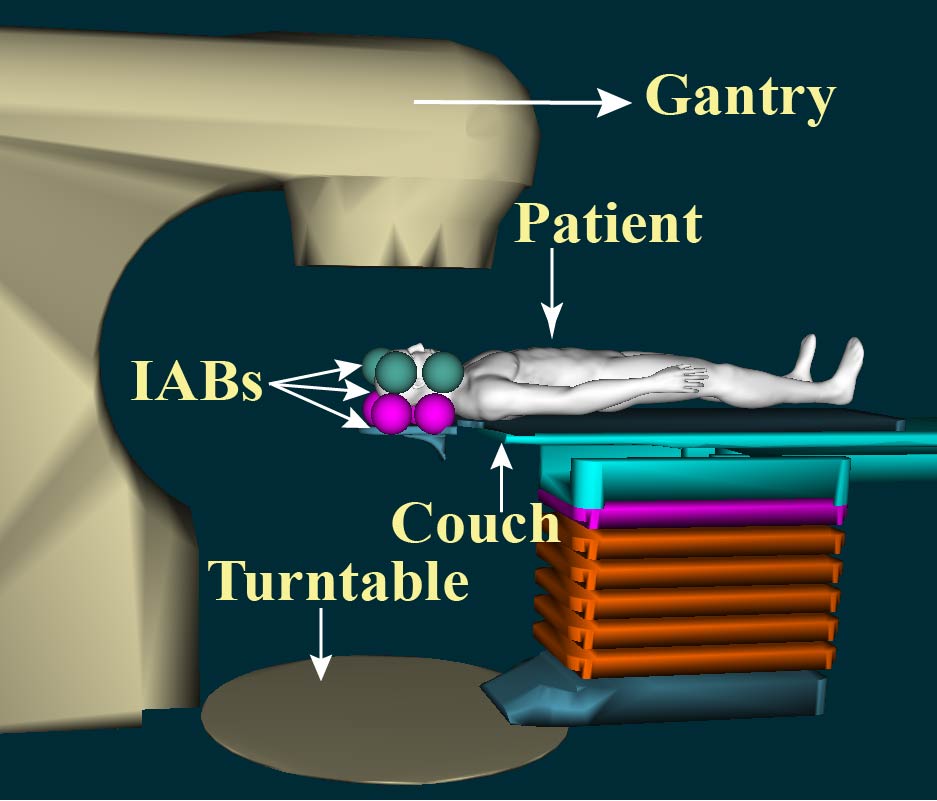}
		\label{fig:sofa_setup_a}
	\end{minipage}
	\begin{minipage}[b]{.24\textwidth}
		\includegraphics[width=\textwidth,height=.9\textwidth]{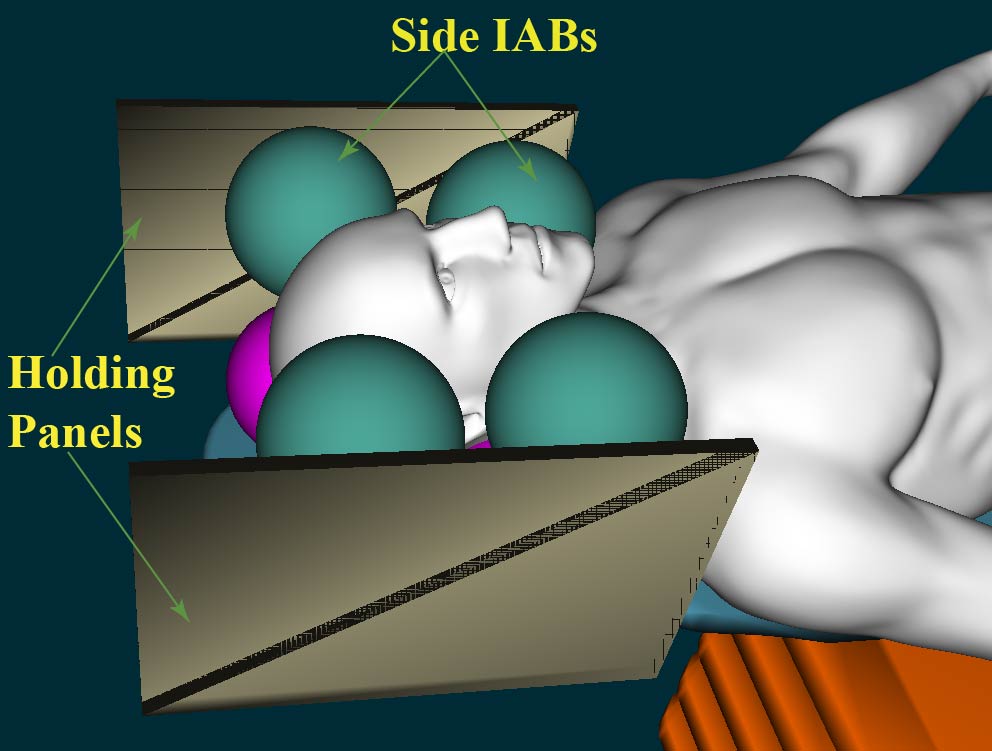} %
		\label{subfig:sofa_setup_b}
	\end{minipage}
	\caption{\textbf{Left}. Gantry, Turntable, Patient and IABs around the patient's H\&N Region (Panel removed for clarity). 
		\textbf{Right}. Close-up setup view with holding PVC foam panels  in the SOFA framework~\cite{SOFA} [Not drawn to scale].
	}
	\label{fig:sofa_setup}
\end{figure}
In this sentiment, we choose eight IABs around the patient's head region as illustrated in \autoref{fig:sofa_setup}. The IABs are held in place around the head by a  low-temperature rigid PVC foam insulation sheet,  encased in carbon fiber to prevent radiation beam attenuation. Velcro stickers (not shown) hold the IABs in place. 

The freedoms provided by each IAB within the setup in \autoref{fig:sofa_setup}b are described as follows: the side actuators correct head motion along the \textit{left-right} axis of the head anatomy, including the yaw and roll motions, while the base IABs correct the head motion along the \textit{anterior-posterior} axis~\cite[Ch. 2]{HuntBook1977}. 
This arrangement offers prehensile manipulation via sensorless motion manipulation strategies within \eg the vision-based sensor plan used in SRS or IMRT. By this, we mean the mechanical interactions of pushing or releasing by the IABs may be harnessed to further improve head manipulation robustness~\cite{Canny94RISC, Mason88, Mason01Book}. 
\begin{figure}[tb!]
	\centering
	\includegraphics[width=.8\columnwidth]{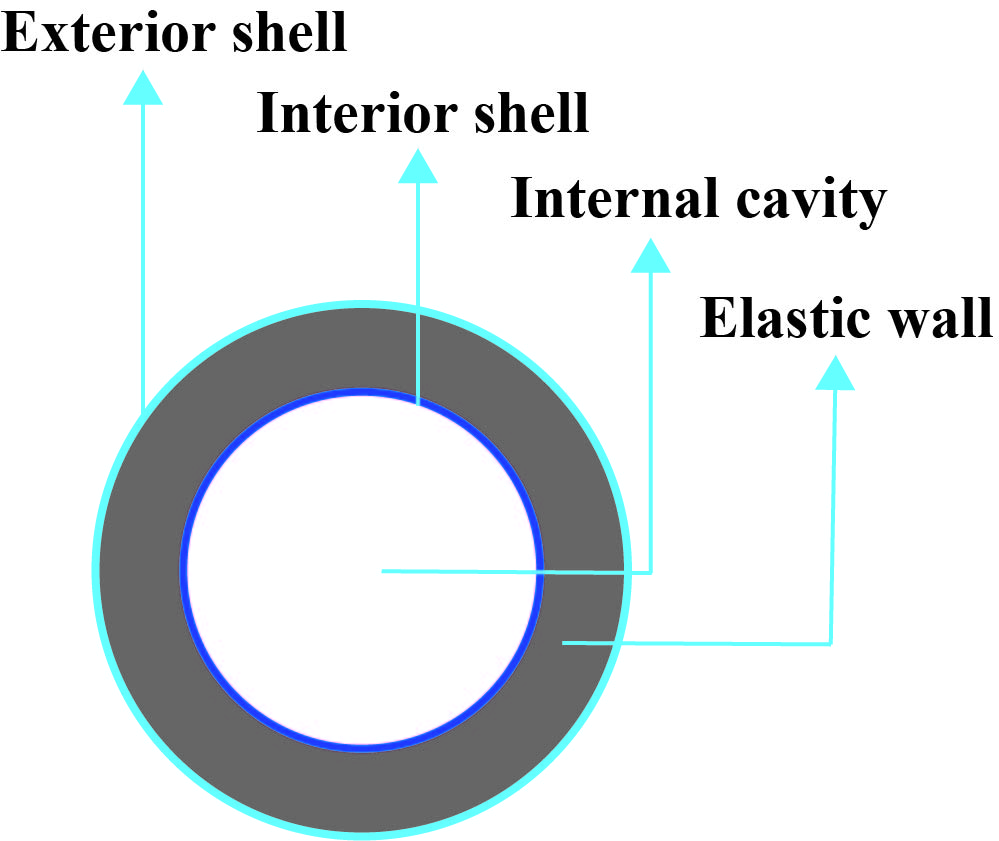}
	\caption{Concentric circular shells around IAB's internal cavity [Not drawn to scale].}
	\label{fig:shells}
\end{figure}
The IABs have an internal cavity that is surrounded by two elastic shells. The shells have radii $2.75 \pm 0.5 cm$ and $3.0 \pm 0.5cm$ respectively and they have constant volume within their wall. The wall thickness is $0.25cm$ in the reference configuration, to accommodate flex and shrinking from repeated use as well as exhibit enough tensile strength that can move the patient's head whilst preserving its compliant properties. The geometry of an IAB is shown in ~\autoref{fig:shells}. The outer shell encapsulates the inner shell so that deformation follows a local volume preservation principle between configuration changes~\cite{OgdenBook}. Deformation is achieved by supplying or removing compressed air from the internal IAB cavity. This fabrication procedure and hardware experiments will be described in a future publication.

%% file: sections/analysis.tex
\section{Deformation Analysis of a Spherical IAB}
\label{sec:deform}

Our overarching assumption is that volume does not change locally during deformation at a configuration $\conf(t)$ at time $t$. We work from a continuum mechanical framework, considering only final configurations for the soft robot; we thus drop the explicit dependence of a configuration on time and write it as $\conf$. We refer readers to background reading materials in~\cite{OgdenBook, Holzapfel2000} and \cite{GentBook}. 
We conclude this section by solving the boundary value problem under the assumption \textit{incompressibility} of the IAB rubber material.

\subsection{Deformation Invariants}
Under the action of applied forces, the IAB's deformation is governed by a stored energy function, $\strain$, which captures the physical properties of the material~\cite{Rivlin1950}. We choose two invariants, $I_1, \text{ and }\, I_2$, described by the principal extension ratios, $\lambda_r, \lambda_\phi, \lambda_\theta$  defined as
\begin{align}
I_1 = \lambda_r^2 + \lambda_\phi^2 + \lambda_\theta^2, \, \text{ and } \quad I_2 =  \lambda_r^{-2} + \lambda_\phi^{-2} + \lambda_\theta^{-2}.
\label{eq:invariants}
\end{align}
\noindent Under the incompressibility assumptions of the IAB material,  it follows that   $\lambda_r\lambda_\phi\lambda_\theta=1$~\cite{Treloar1975}. 
\begin{figure}[!tb]
	\centering
	\begin{minipage}[b]{0.23\textwidth}
		\includegraphics[width=\textwidth]{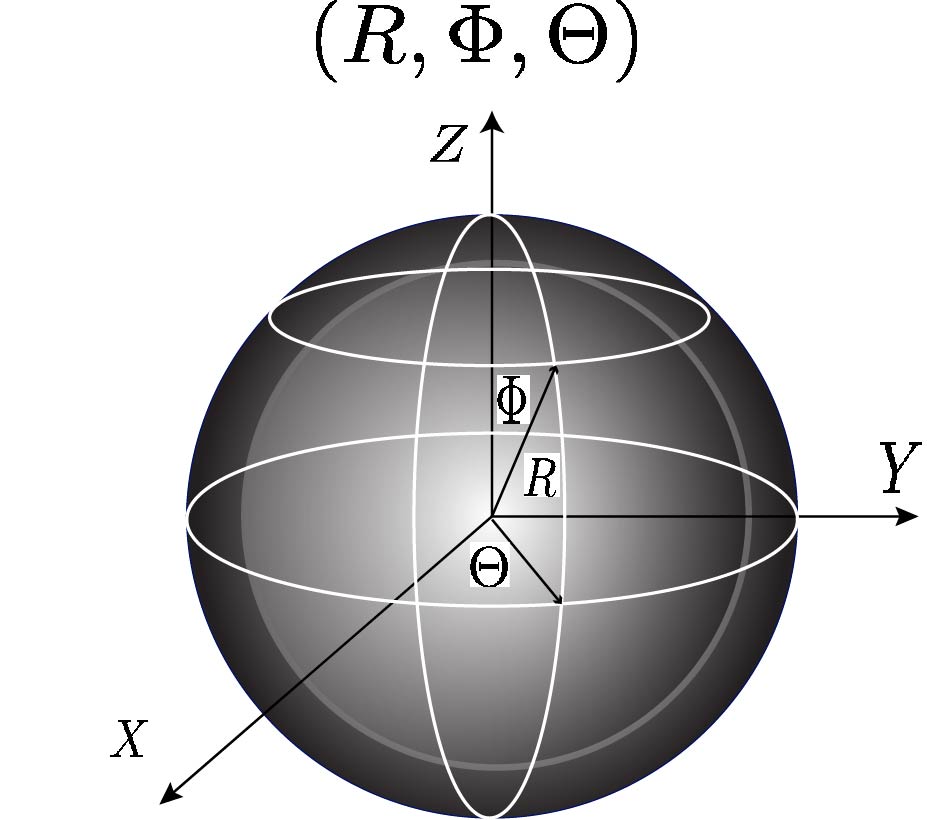}
		\vspace{0.1em}
		\subcaption{Reference configuration}
	\end{minipage}
	\begin{minipage}[b]{0.25\textwidth}
		\includegraphics[width=\textwidth]{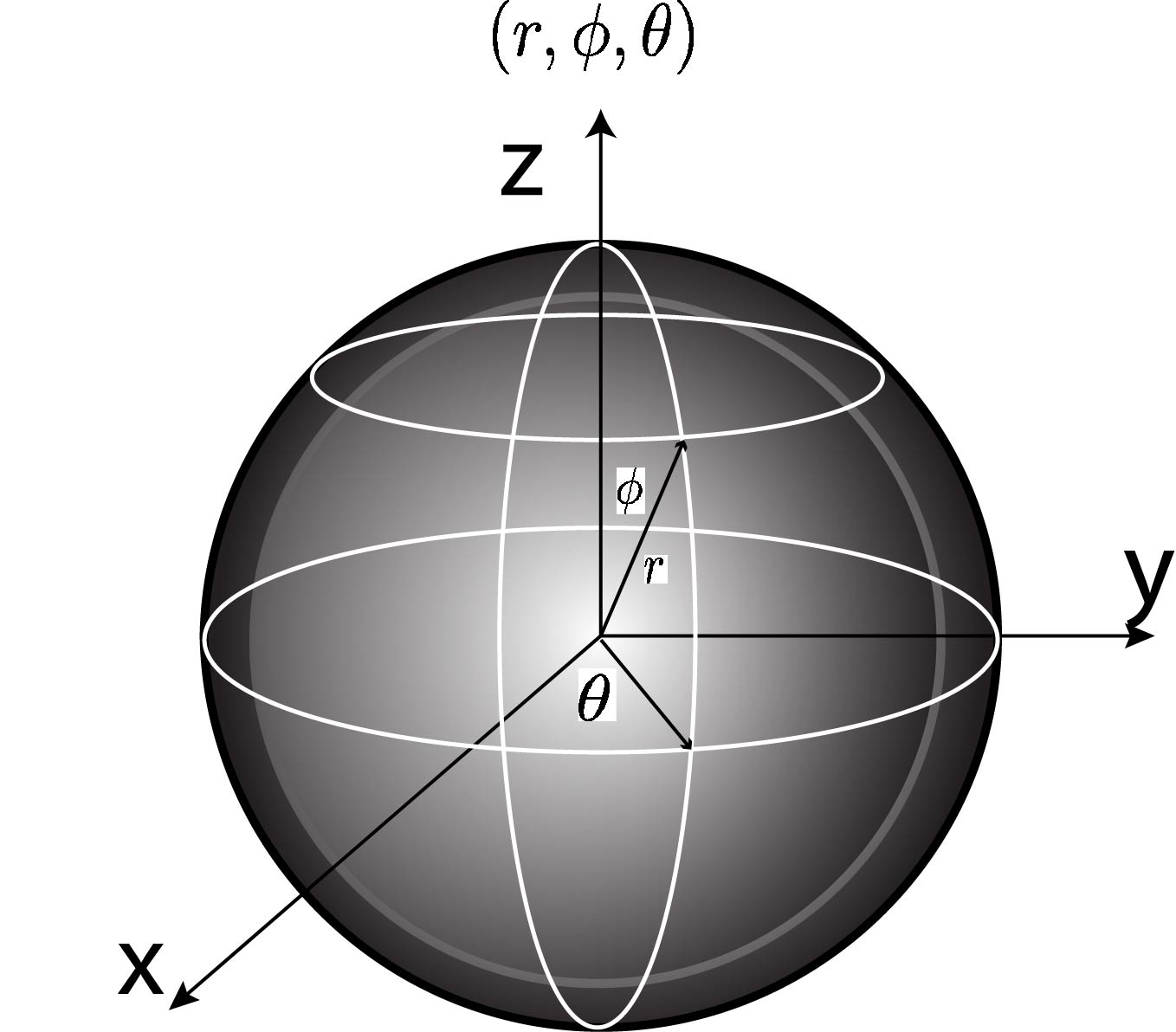}
		\vspace{0.1em}
		\subcaption{Current configuration}
	\end{minipage}
	\caption{IAB configurations in spherical polar coordinates.}
	\label{fig:spherical_coords}
\end{figure}
In spherical coordinates, the change in polar/azimuth angles as well as radii from the reference to current configurations are as illustrated in \autoref{fig:spherical_coords}. 
Forces that produce deformations are derived using the strain energy-invariants relationship, \ie, $\frac{\partial \strain}{\partial I_1}$ and $\frac{\partial \strain}{\partial I_2}$.

\subsection{Analysis of Strain Deformations}
Suppose a particle on the IAB material surface in the reference configuration has coordinates $(R, \Phi, \Theta)$ defined in spherical polar  coordinates (see \autoref{fig:spherical_coords}), where $R$ represents the radial distance of the particle from a fixed origin, $\Theta$ is the azimuth angle on a reference plane through the origin and orthogonal to the polar angle, $\Phi$. Denote the  internal and external radii as $R_i \text{ and } R_o$ respectively. We define the following constraints,
\begin{align}
R_i \le R \le R_o, \quad 0 \le \Theta \le 2\pi, \quad 0 \le \Phi \le \pi.
\label{eq:polar_coordinates_ref}
\end{align}
%
%
Now, suppose that the IAB undergoes deformation upon pressurization of its internal cavity: arbitrary points $A$ and $A'$ in the reference configuration become $Q$ and $Q'$ in the current configuration. Let the \textit{material element} (or fiber) vector that connects  points $A$ and $A'$ be $a = a_R \basis_r + a_\Theta \basis_\Theta + a_\Phi \basis_\Phi$, where $\basis_R, \basis_\Theta$, and $\basis_\Phi$ are respectively the basis vectors for polar directions $R, \Theta$, and $\Phi$ such that its axial length stretches \textit{uniformly} by an amount $\lambda_z = \frac{r}{R}$. 
If  spherical symmetry is maintained during deformation, we have  the following constraints in the current configuration
\begin{align}
r_i \le r \le r_o, \quad 0 \le \theta \le 2\pi,  \quad 0 \le \phi \le \pi.
\label{eq:polar_coordinates_curr}
\end{align}
We define radial vectors $\textbf{R}$ and $\textbf{r}$ in spherical coordinates as,
\begin{align}
\textbf{R} = \begin{bmatrix}
R \, \cos \Theta \, \sin \Phi, \\ R \, \sin \Theta \, \sin \Phi, \\ R \, \cos  \Phi
\end{bmatrix} \quad \text{and} \quad 	
\textbf{r} = \begin{bmatrix}
r \, \cos \theta \, \sin \phi, \\ r \, \sin \theta \, \sin \phi, \\ r \, \cos  \phi
\end{bmatrix}.
\label{eq:spherical-position}
\end{align}
%
%
%
%
%
The material volume $\frac{4}{3}\pi\left(R^3 - R_i^3\right)$ contained between spherical shells of radii $R$ and $R_i$ remains constant throughout deformation, being equal in volume to $\frac{4}{3}\pi\left(r^3 - r_i^3\right)$  so that 
\begin{align}
\dfrac{4}{3}\pi\left(R^3 - R_i^3\right) &= \dfrac{4}{3}\pi\left(r^3 - r_i^3\right) \nonumber \\
\text{ or } r^3 &= R^3 + r_i^3 - R_i^3.
 \label{eq:volume_preservation}
\end{align}
The homogeneous deformation between the two configurations 
imply that
\begin{align}
r^3 = R^3 + r_i^3 - R_i^3, \quad \theta = \Theta, \quad \phi = \Phi ,
\label{eq:spherical_transformation}
\end{align}
where the coordinates obey the constraints of equations \eqref{eq:polar_coordinates_ref} and \eqref{eq:polar_coordinates_curr}. 
The Mooney-Rivlin strain energy for small deformations as a function of the strain invariants of \eqref{eq:invariants}, is,
\begin{align}
\strain^\prime = C_1(I_1-3) + C_2(I_2-3),
\label{eq:Mooney}
\end{align}
where  $C_1$ and $C_2$ are appropriate choices for the IAB material moduli. The Mooney form \eqref{eq:Mooney} has been shown to be valid even for large elastic deformations, provided that the elastic materials exhibit incompressibility and are isotropic in their reference configurations~\cite{Mooney1940}. For mathematical scaling purposes that will  become apparent, we rewrite  \eqref{eq:Mooney} as $W= \frac{1}{2}W^\prime$ so that
%
\begin{align}
\strain = \frac{1}{2}C_1(I_1-3) + \frac{1}{2}C_2(I_2-3).
\label{eq:Mooney_scaled}
\end{align}
The deformation gradient tensor in spherical-polar coordinates can be verified to be
\begin{align}
\deformationgrad &= \lambda_r \basis_r \otimes \basis_R + \lambda_\phi \basis_\phi \otimes \basis_\Phi + \lambda_\theta \basis_\theta \otimes \basis_\Theta \nonumber \\
&= \frac{R^2}{r^2}  \basis_r \otimes \basis_R +  \frac{r}{R} \basis_\phi \otimes \basis_\Phi + \frac{r}{R} \basis_\theta \otimes \basis_\Theta,	
\label{eq:deformation_grad}
\end{align}
where $\otimes$ denotes the dyadic tensor product. It can be  verified that the radial stretch is $\lambda_r = \frac{R^2}{r^2}$. 
The principal stretches along the azimuthal and zenith axes imply that $\lambda_\theta = \lambda_\phi$. Since for an isochoric deformation, $ \lambda_r \cdot \lambda_\theta \cdot \lambda_\phi =1$, the principal extension ratios are
\begin{align}
\lambda_r =  \frac{R^2}{r^2}; \lambda_\theta = \lambda_\phi  = \frac{r}{R}.
\end{align}
The invariant equations, in spherical-polar coordinates, are therefore a function of the right Cauchy-Green and finger deformation tensors~\cite{IUPAC} \ie, 
\begin{align}
	I_1 &= \trace{\rightcauchy} =  \dfrac{R^4}{r^4} + \dfrac{2 \, r^2}{R^2}, \,\,
	I_2 = \textbf{tr}\left(\rightcauchy^{-1}\right) = \dfrac{r^4}{R^4} + \dfrac{2 \, R^2}{r^2}
	\label{eq:invariants_polar}
\end{align}
where, $\rightcauchy=\deformationgrad^T \deformationgrad$ and $\leftcauchy = \deformationgrad \deformationgrad^T$ are the right and left Cauchy-Green tensors respectively. 
\subsection{Stress Laws and Constitutive Equations}
We are concerned with the magnitudes of the differential stress on the IAB skin from a mechanical point of view. \textbf{We do not rely on finite element methods in this work} but rather take the whole material as a single continuum structure with elastic properties. Since the IAB deforms at ambient temperature, we take thermodynamic properties such as temperature and entropy to have negligible contribution.  The IAB material stress response, $\materialresponse$, at any point on the IAB's boundary at time $t$ determines the Cauchy stress, $\cauchystress$, as well as the history of the motion up to and at the time $t$~\cite{OgdenBook}. 
\begin{figure}[tb!]
	\centering
	\begin{minipage}[b]{0.5\textwidth}
		\includegraphics[width=.8\textwidth]{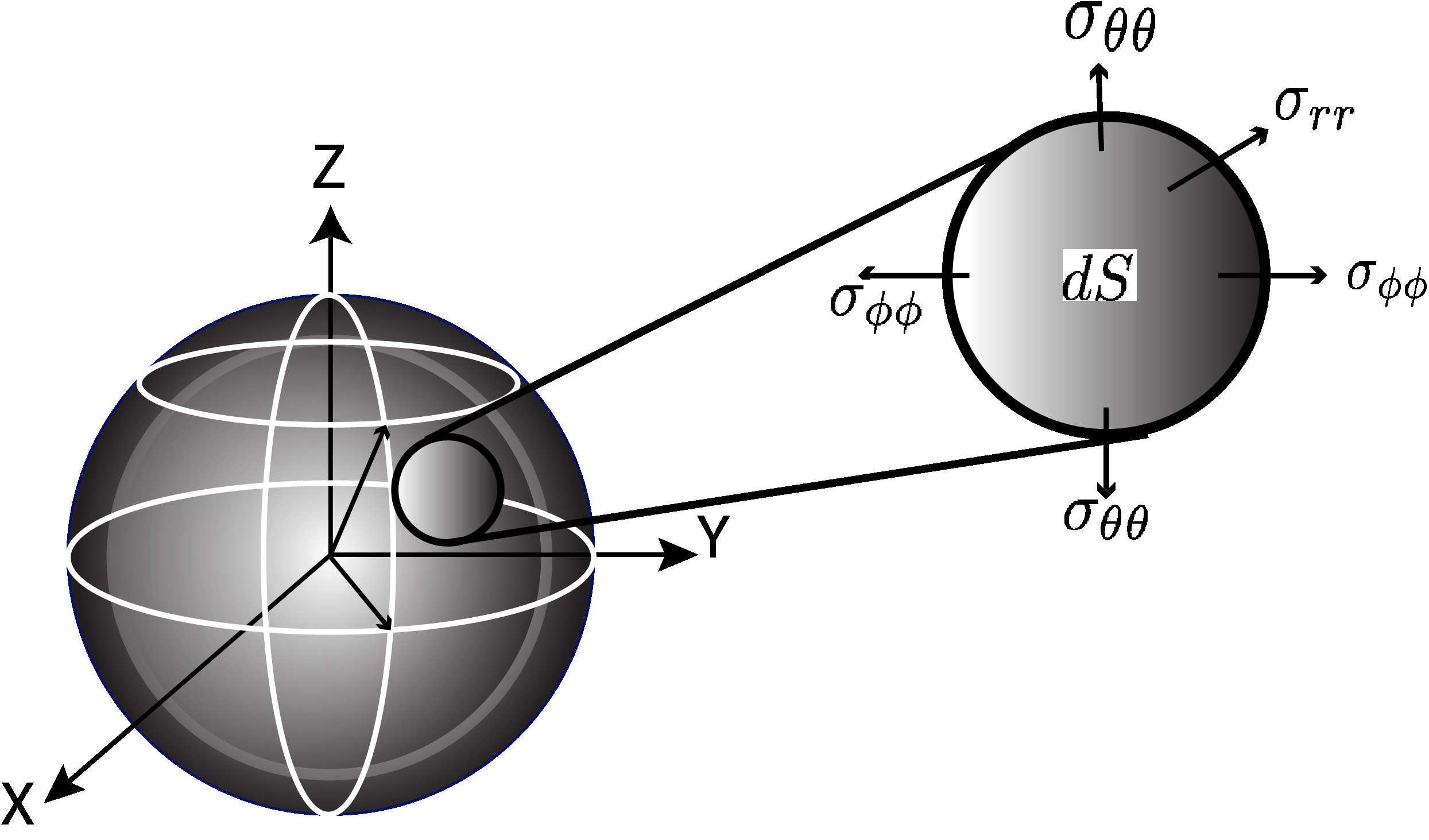}
	\end{minipage}
	%
	\caption{Stress distribution on the IAB's differential surface, $dS$.}
	\label{fig:fibres}
\end{figure}
The constitutive relation for the nominal stress deformation for an elastic IAB material is given by
\begin{align}
\cauchystress = \materialresponse(\deformationgrad) + q \deformationgrad \dfrac{\partial \bm{\Lambda}}{\partial \deformationgrad}(\deformationgrad),
\label{eq:cauchy_stress1}
\end{align}
where $\materialresponse$ is a functional with respect to the configuration $\conf_t$, $q$ acts as a Lagrange multiplier, and $\bm{\Lambda}$ denotes the internal (incompressibility) constraints of the IAB system. For an incompressible material, the indeterminate Lagrange multiplier becomes the hydrostatic pressure \ie $q=-p$~\cite{Holzapfel2000}. The incompressibility of the IAB material properties imply that $\bm{\Lambda} \equiv \text{det }\deformationgrad - 1$. 
Evaluating the partial derivative of $\bm{\Lambda}(\deformationgrad)$ with respect to $\deformationgrad$ and substituting $-p$ for $q$ in  \eqref{eq:cauchy_stress1}, we can verify that
\begin{align}
\cauchystress &= 
\materialresponse(\deformationgrad) - p \identity 
\label{eq:cauchy_stress2}
\end{align}
following the isochoric assumption \ie, $\text{det}(\deformationgrad) = 1$. 
In terms of the stored strain energy, we find that 
\begin{align}
\cauchystress 
= \dfrac{\partial W}{\partial \deformationgrad} \deformationgrad^T - p\identity
\label{eq:cauchy_stress}
\end{align}
%
%
where $\identity$ is the identity tensor and $p$ represents an arbitrary hydrostatic pressure. 
It follows that  the constitutive law that governs the Cauchy stress tensor is 
\begin{align}
\cauchystress &= \dfrac{\partial W}{\partial \text{I}_1}\cdot \dfrac{\partial \text{I}_1}{\partial \deformationgrad}\deformationgrad^T + \dfrac{\partial W}{\partial \text{I}_2}\cdot \dfrac{\partial \text{I}_2}{\partial \deformationgrad}\deformationgrad^T - p\identity \nonumber \\
&= \frac{1}{2} C_1 \dfrac{\partial \textbf{tr}\left(\deformationgrad \deformationgrad^T\right)}{\partial \deformationgrad}\deformationgrad^T + \frac{1}{2} C_2\dfrac{\partial \trace{\left[\deformationgrad^T \,  \deformationgrad\right]^{-1}}}{\partial \deformationgrad}\deformationgrad^T  - p\,\identity \nonumber \\
&= \frac{1}{2} C_1 \left(2 . \deformationgrad \deformationgrad^T\right)+ \frac{1}{2} C_2 \left(-2\deformationgrad (\deformationgrad^T \,  \deformationgrad)^{-2} \right) \deformationgrad^T  - p\,\identity \nonumber \\
&=  C_1 \deformationgrad \deformationgrad^T - C_2 \left(\deformationgrad^T \deformationgrad\right)^{-1}- p \identity  \nonumber \\
\cauchystress  &= C_1 \leftcauchy - C_2 \rightcauchy^{-1}  - p\identity, 
\label{eq:stress_constitutive}
\end{align}
from which we can write the normal stress components as
\begin{subequations}
	\begin{align}
	\stresscomp_{rr} &= -p + C_1 \dfrac{R^4}{r^4} - C_2 \dfrac{r^4}{R^4} \\
	\stresscomp_{\theta \theta} &= \stresscomp_{\phi \phi} = -p + C_1 \dfrac{r^2}{R^2} - C_2 \dfrac{R^2}{r^2}. 
	\end{align}
	\label{eq:stress_compos}
\end{subequations}
A visualization of the component stresses of \eqref{eq:cauchy_stress} on the outer shells of the IAB material is illustrated in \autoref{fig:fibres}. 
\subsection{IAB Boundary Value Problem}
\label{subsec:contact-free-bvp}
Here, we analyze the stress and internal pressure of the IAB at equilibrium.  
Consider the IAB with boundary conditions
\begin{align}
\stresscomp_{rr}|_{R=R_o} = -P_\text{atm}, \quad \stresscomp_{rr}|_{R=R_i} = -P_\text{atm} - P
\label{eq:boundary_conds}
\end{align}
where $P_\text{atm}$ is the atmospheric pressure and $P>0$ is the internal pressure exerted on the walls of the IAB above $P_\text{atm}$ \ie, $P > P_\text{atm}$. Suppose that the IAB stress components satisfy hydrostatic equilibrium, the equilibrium equations for the body force $\bm{b}'s$ physical component vectors, $b_r, b_\theta, b_\phi$ are
\begin{subequations}
	\begin{align}
	-b_r &= \frac{1}{r^2}\frac{\partial r^2 \stresscomp_{rr}}{\partial r} + \frac{1}{r \sin\phi}\frac{\partial \sin \phi \stresscomp_{r \phi}}{\partial \phi} + \frac{1}{r \sin\phi}\frac{\partial \stresscomp_{r \theta}}{\partial \theta} \nonumber \\
	& -  \frac{1}{r}(\stresscomp_{\theta\theta}+ \stresscomp_{\phi \phi})
	\label{eq:polar_coord_a}
	\end{align}
	\begin{align}
	-b_\phi &= \frac{1}{r^3}\frac{\partial r^3 \stresscomp_{r\phi}}{\partial r} + \frac{1}{r \sin\phi}\frac{\partial \sin \phi \stresscomp_{\phi \phi}}{\partial \phi} + \frac{1}{r \sin\phi}\frac{\partial \stresscomp_{\theta \phi}}{\partial \theta} \nonumber \\
	&-  \frac{\cot \phi}{r}(\stresscomp_{\theta\theta})
	\label{eq:polar_coord_b}
	\end{align}
	\begin{align}
	-b_\theta &= \frac{1}{r^3}\frac{\partial r^3 \stresscomp_{\theta r}}{\partial r} + \frac{1}{r \sin^2\phi}\frac{\partial \sin^2 \phi \stresscomp_{\theta \phi}}{\partial \phi} + \frac{1}{r \sin\phi}\frac{\partial \stresscomp_{\theta \theta}}{\partial \theta}
	\label{eq:polar_coords_c}
	\end{align}
	\label{eq:polar_coords}
\end{subequations}
(see ~\cite{FungandTong}). From the equation of balance of linear momentum (\textit{Cauchy's first law of motion}), we have that
\begin{align}
\text{div } \cauchystress^T  + \rho \textbf{b} = \rho \dot{\textbf{v}}
\label{eq:cauchy_law1}
\end{align}
where $\rho$ is the IAB body mass density, $\text{div}$ is the divergence operator, and $\textbf{v}(\particle, t) = \dot{\conf}_t(\bm{X})$ is the velocity gradient. Owing to the incompressibility assumption, we remark in passing that the mass density is uniform throughout the body of the IAB material. When the IAB is at rest, $\dot{\textbf{v}}_t(\particle)=0 \, \forall \, \particle \in \mc{B}$ such that equation \eqref{eq:cauchy_law1} loses its dependence on time. The assumed regularity of the IAB in the reference configuration thus leads to the steady state conditions for Cauchy's first equation;  the stress field $\cauchystress$ becomes \textit{self-equilibrated} by virtue of the spatial divergence and the symmetric properties of the stress tensor, so that we have 
\begin{align}
\textbf{div } \cauchystress = 0.
\label{eq:divergence}
\end{align}
%
%
Equation \ref{eq:divergence} is satisfied if the hydrostatic pressure $p$  in \eqref{eq:stress_constitutive} is independent of $\theta$ and $\phi$. Therefore, we are left with \eqref{eq:polar_coord_a} so that we have 
\begin{align}
\frac{1}{r}\frac{\partial}{\partial r}(r^2 \stresscomp_{rr}) = (\stresscomp_{\theta\theta}+ \stresscomp_{\phi \phi}).
\end{align}
Expanding, we find that
\begin{align}
\frac{1}{r}\left[r^2\frac{\partial \stresscomp_{rr}}{\partial r} + \stresscomp_{rr}\frac{\partial (r^2)}{\partial r}  \right] &= (\stresscomp_{\theta\theta}+ \stresscomp_{\phi \phi}) \nonumber \\
%
%
r\frac{\partial \stresscomp_{rr}}{\partial r} &= \stresscomp_{\theta\theta}+ \stresscomp_{\phi \phi} - 2 \stresscomp_{rr}  \\
\frac{\partial \stresscomp_{rr}}{\partial r}  &= \frac{1}{r}(\stresscomp_{\theta\theta}+ \stresscomp_{\phi \phi} - 2 \stresscomp_{rr}). 
\end{align}
Integrating the above equation in the variable $r$, and taking $\stresscomp_{rr}(r_\circ)=0$,  we find that
\begin{align}
\stresscomp_{rr}(r) &= -\int_{r_i}^{r_\circ} \frac{1}{r}(\stresscomp_{\theta\theta} + \stresscomp_{\phi\phi}-2\stresscomp_{rr}) dr, 
\nonumber  \\
&=  - \int_{r_i}^{r_\circ} \left[2 C_1\left(\frac{r}{R^2}-\frac{R^4}{r^{5}}\right)+2C_2\left(\frac{r^{3}}{R^4} - \frac{R^2}{r^{3}}\right)\right]  dr.
\end{align}
The above relation gives the radial stress in the current configuration. Suppose we are in the current configuration and we desire to revert to the reference configuration, we may carry out a change of variables from $r$ to $R$ as follows,
\begin{align}
&\stresscomp_{rr}(R) = -\int_{R_i}^{R_\circ} \frac{1}{r}(\stresscomp_{\theta\theta} + \stresscomp_{\phi\phi}- 2\stresscomp_{rr}) \dfrac{dr}{dR} dR, 
\nonumber \\
%
%
&=     -\int_{R_i}^{R_\circ} \left[2 C_1\left(\frac{1}{r}-\frac{R^6}{r^{7}}\right)-2C_2\left(\frac{R^{4}}{r^5} - \frac{r}{R^{2}}\right) \right]  dR.
\label{eq:boundary_integrand}
\end{align}
%
In the same vein, using the boundary condition of \eqref{eq:boundary_conds}$|_2$ and taking the ambient pressure $P_{\text{atm}} = 0$, we find that the internal pressure $P = -\stresscomp_{rr}(r)$ so that
%
\begin{align}
&P(r)  =  \int_{r_i}^{r_\circ} \left[2 C_1\left(\frac{r}{R^2}-\frac{R^4}{r^{5}}\right)+2C_2\left(\frac{r^{3}}{R^4} - \frac{R^2}{r^{3}}\right)\right]  dr \nonumber  \\
&P(R)\equiv \int_{R_i}^{R_\circ} \left[2 C_1\left(\frac{1}{r}-\frac{R^6}{r^{7}}\right)-2C_2\left(\frac{R^{4}}{r^5} - \frac{r}{R^{2}}\right) \right] dR.
\label{eq:internal_pressure}
\end{align}
Equations \eqref{eq:boundary_integrand} and \eqref{eq:internal_pressure} completely determine the  inverse  kinematics of the IAB material: given a desired expansion or compression of the IAB walls, it calculates the internal pressurization or stress tensor necessary to achieve such deformation. Under the incompressibility of the IAB material properties we have 
\begin{align}
{r}^3 = R^3 + r_i^3 - R_i^3, \text{ and } {r_\circ}^3 = R_\circ^3 + r_i^3 - R_i^3.
\end{align}

%% file: sections/simulation.tex
\section{Simulation Results}
\label{sec:simu}
\begin{figure}[tb!]
\centering
\begin{minipage}[tb]{0.4850\textwidth}
	\includegraphics[width=\textwidth]{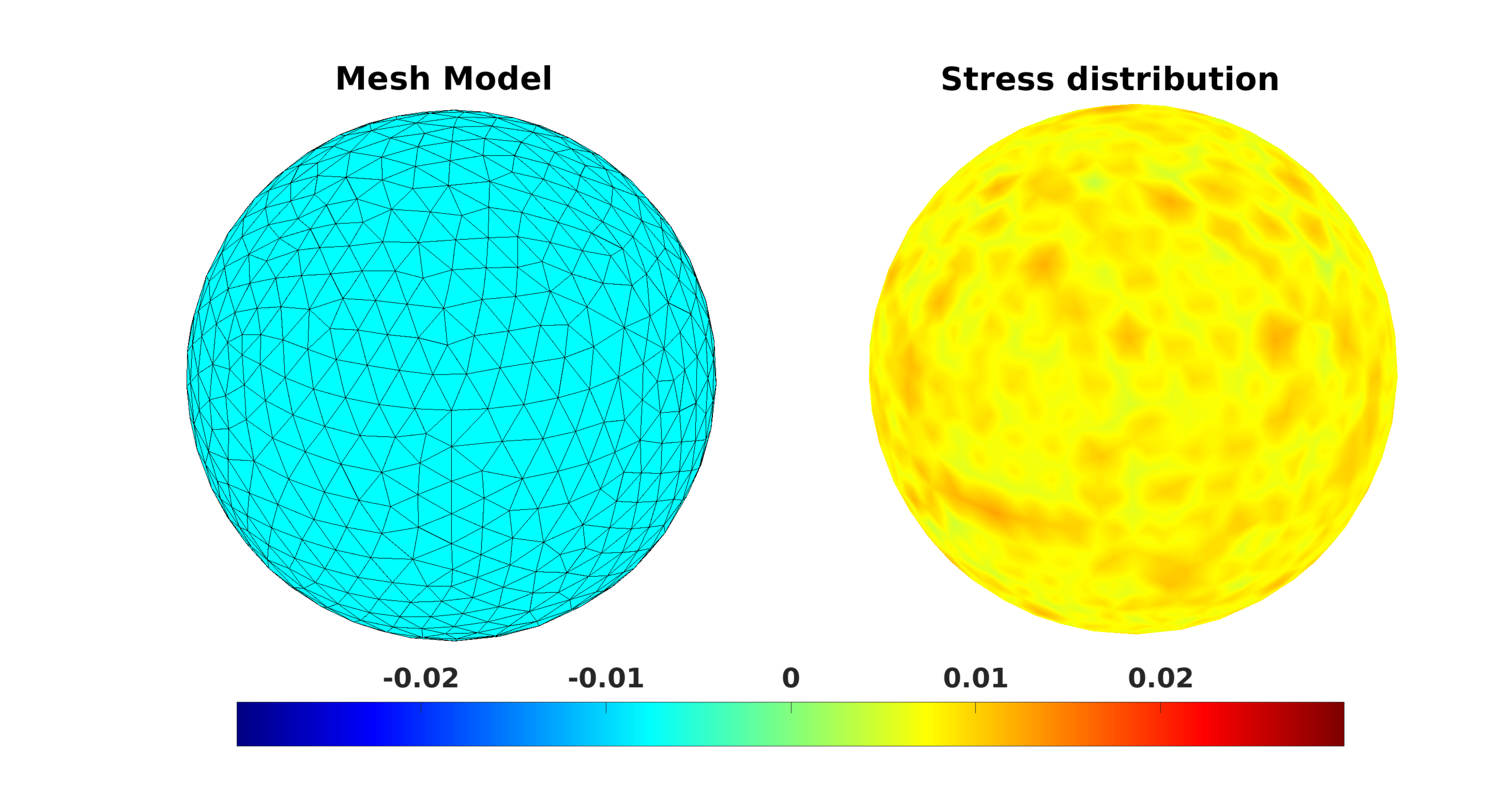}
	\subcaption{\textbf{Left}: Mesh model. \textbf{Right}: Stress distribution on outer skin.}
\end{minipage}
\hfill
\begin{minipage}[tb]{.485\textwidth}
	\includegraphics[width=\textwidth]{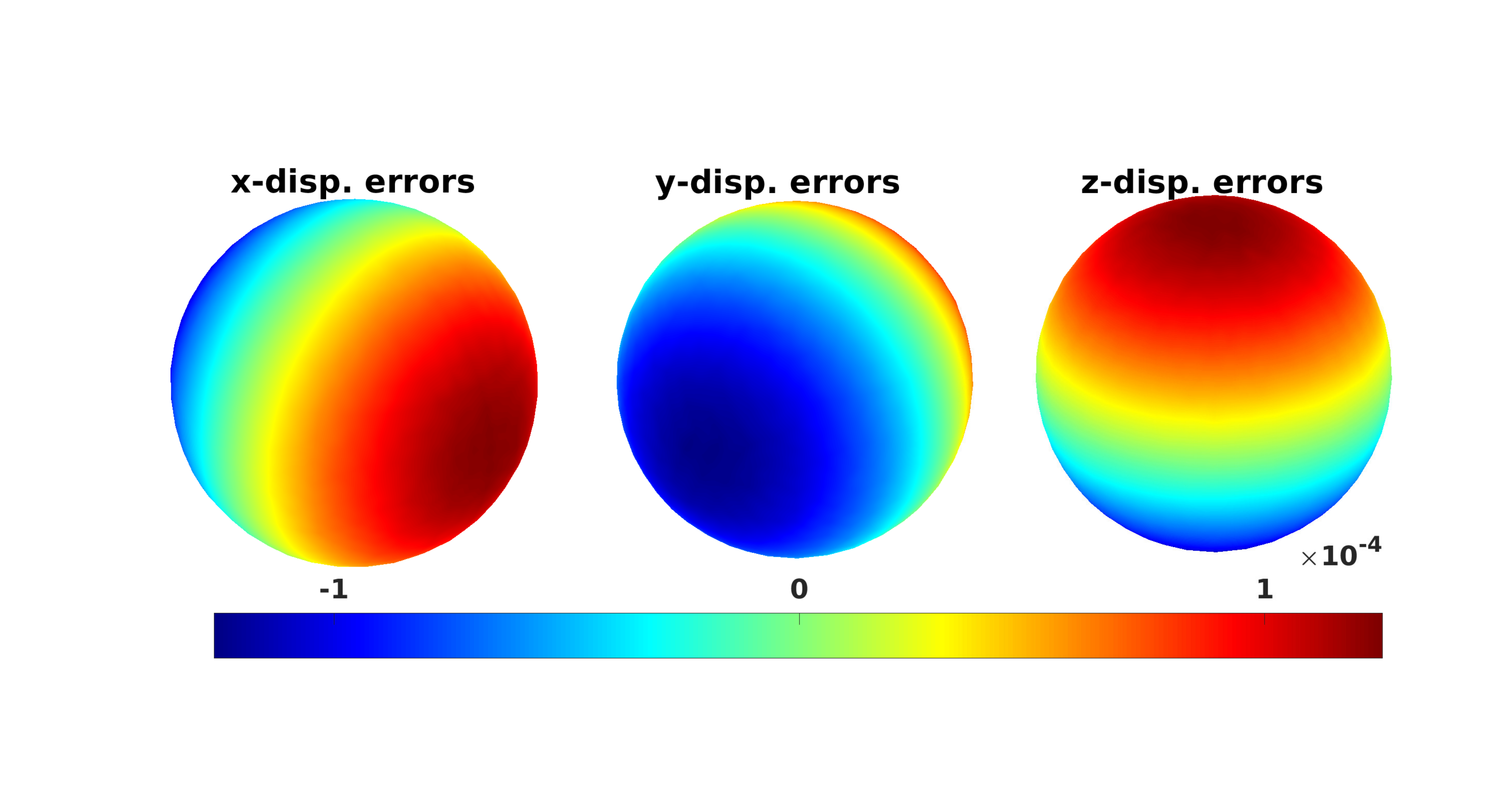}
	\subcaption{Displacement errors along  $x,y,z$  coordinates.} 
\end{minipage}
\vspace{0.5em}
\begin{tabular}{|c|c|c|c|c||c|c|c|c|}
	\hline 
	\multicolumn{5}{|c||}{Inputs}
	&  \multicolumn{3}{|c|}{Outputs} \\ \cline{1-8}$C_1$ &  $C_2$ & $R_i$ &  $r_i $  & $R_\circ$ & $r_\circ$ & $P $  & $\Delta V$  \\
	\hline \rule[-2ex]{0pt}{5.5ex}  $1.1e4$ &  $2.2e4$ & $.025$ & $.03$ & $.03$  & $.028$ & $-.34$ & $\approxeq 0$    \\
	\hline
\end{tabular} 
\caption{Volumetric Deformation (Compression).}
\label{fig:deform_compress_1}
\end{figure}

We conduct simulations under volumetric deformation with different shell properties (stated in the tables of \autoref{fig:deform_extend_1} and \ref{fig:deform_compress_1}). We fixed both reference configuration radii and choose appropriate  volumetric moduli for the IAB shells (with $C_1$ being the material's Young's modulus and $C_2$ its stiffness). By specifying a desired radially symmetric expansion for the inner IAB material, we test the local volume preservation property of \eqref{eq:volume_preservation} and evaluate the resulting displacement of the outer IAB skin, by  applying the computed pressure (by virtue of \eqref{eq:internal_pressure}) between configurations. 

We used the partial differential equations toolbox in Matlab in creating multispheres that correspond to the size synthesis described in \autoref{subsec:num_syn} and in solving the integrands of \eqref{eq:boundary_integrand} and \eqref{eq:internal_pressure}.
The computed soft mesh model of the IAB is shown in the top left of the charts while the stress distribution after the application of the calculated pressure (by virtue of \eqref{eq:internal_pressure}) is shown in the top-right of the figures. We chose a Poisson's ratio of $\approx 0.45$ (since the material shells are made of incompressible rubber materials) and a uniform mass density set at $0.1 kg/m^3$ for the IAB (owing to its volume preserving property).  The radii dimensions  are in $m$, the pressure is given in Pascals unless otherwise stated, $C_1$ and $C_2$ are appropriate material moduli, and $\Delta V$ is the volumetric change between the IAB shells between configurations (given in $m^3$).

\subsection{Volumetric Expansion}
\begin{figure}[tb!]
	\centering
	\begin{minipage}[tb]{0.4850\textwidth}
	\includegraphics[width=\textwidth]{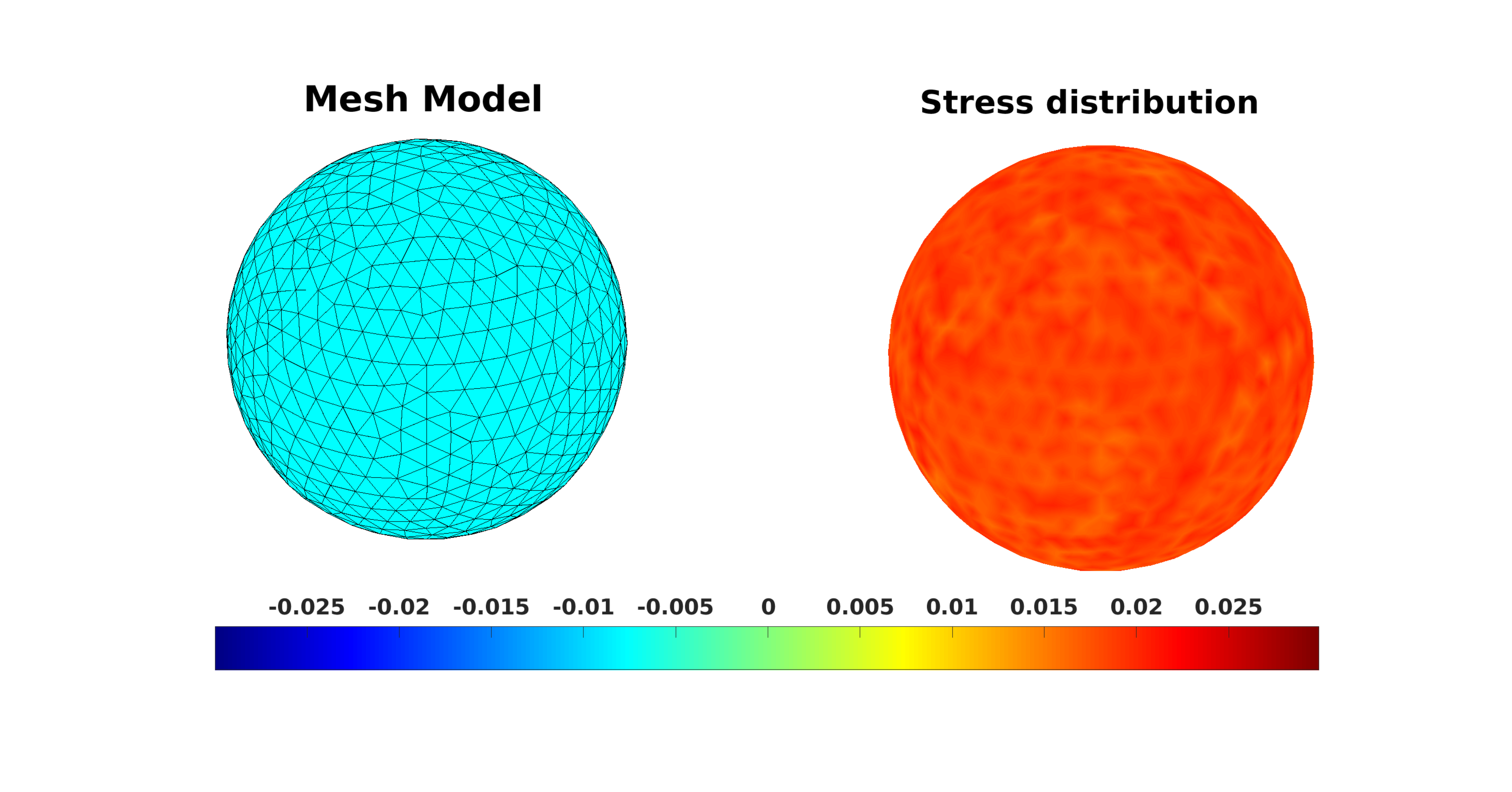}
	\subcaption{\textbf{Left}: Mesh model. \textbf{Right}: Stress distribution on outer skin.}
	\end{minipage}
	\hfill
	\begin{minipage}[tb]{.485\textwidth}
		\includegraphics[width=\textwidth]{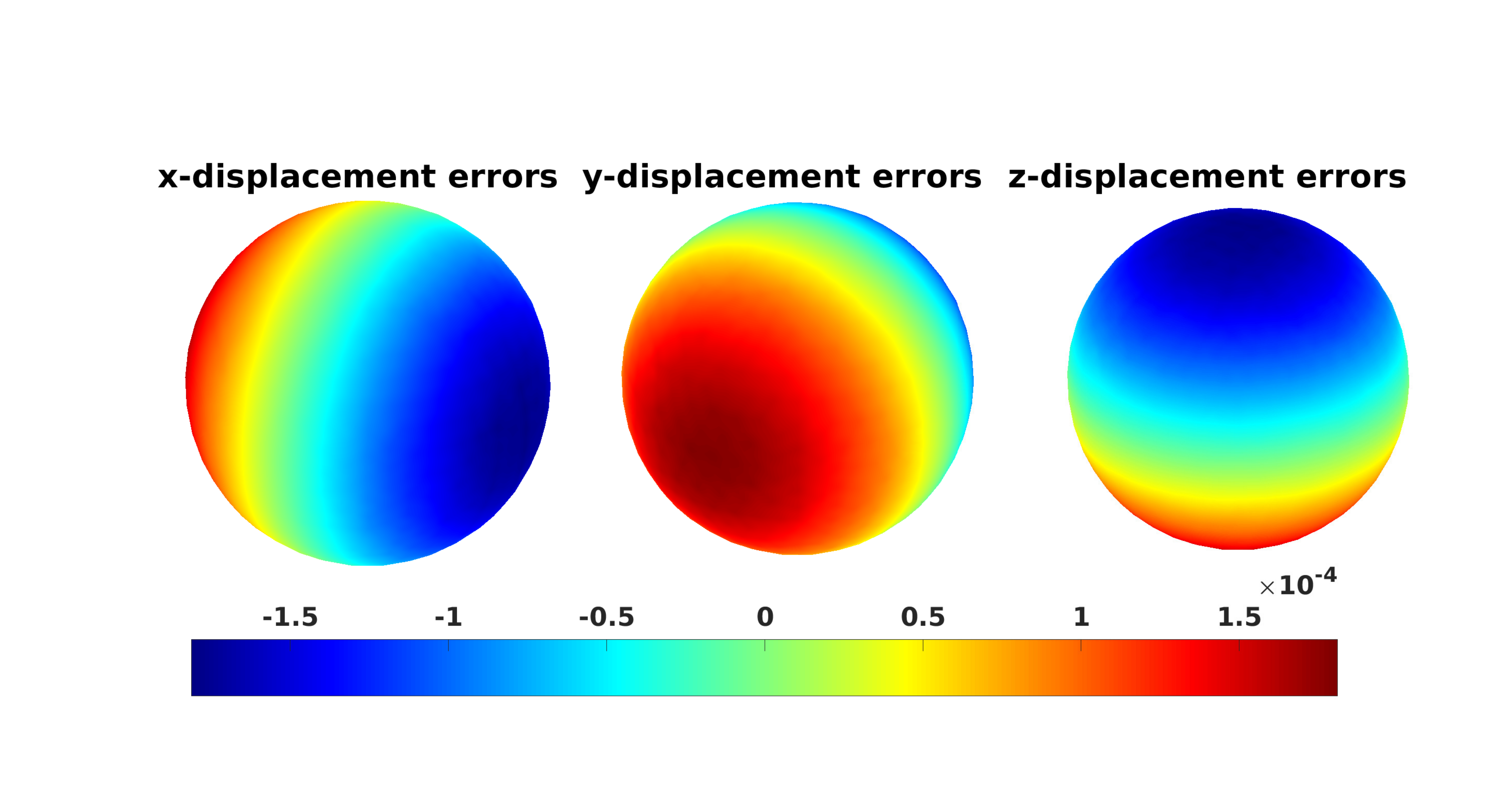} 
		\subcaption{Displacement errors along $x,y,z$ coordinates.} 
	\end{minipage}
	\begin{tabular}{|c|c|c|c|c||c|c|c|c|}
		\hline 
		\multicolumn{5}{|c||}{Inputs}
		&  \multicolumn{3}{|c|}{Outputs} \\ \cline{1-8}
		 $C_1$ &  $C_2$ & $R_i$ &  $r_i $ & $R_\circ$ & $r_\circ$ & $P$  & $\Delta V$  \\
		\hline  $1.1e4$ &  $2.2e4$ & $.027$ & $.03$  & $.03$ & $.033$& $.76$ & $\approxeq 0 $  \\
		\hline
	\end{tabular} 
	\caption{Volumetric Deformation (Expansion).}
	\label{fig:deform_extend_1}
\end{figure}

In \autoref{fig:deform_extend_1}, we test finite elastic deformation of the IAB material shells. The internal and external radii in the reference configuration are $0.027m$ and $0.03m$. This is consistent with the size and shape of the the head of an adult human head from Cadaver studies~\cite{Clauser69, Walker71, Walker73, Dempster55}. The task here is to to achieve a volumetric expansion so that in the current configuration, the internal radius is $0.03m$. By \eqref{eq:spherical_transformation}, we found $r_\circ$ to be $0.033m$. The stress distribution on the skins of the IAB is uniformly distributed based on the single value of the stress in every region of the surface; this signifies an equal amount of stress exertion on the walls of the actuator to achieve a desired deformation. This is confirmed in the bottom part of \autoref{fig:deform_extend_1} where we notice a displacement error of $0.15mm$, precise enough for prehensile motion manipulation of the head as we would require in the enumerated applications for this soft actuator model. 

\subsection{Volumetric Compression}
\autoref{fig:deform_compress_1}
depicts the volumetric compression of the incompressible IAB under the application of the derived internal pressure. For a desired uniform displacement $0.002m$, our results confirm the validity of the stress-strain constitutive laws, as we again notice a displacement error of $\approxeq10^{-4}m$  along the sphere's Cartesian axes in the lower charts of \autoref{fig:deform_compress_1}.  The negative pressure in the table signify that air is being drawn out of the IAB. 
Again, our results show consistency with respect to local volume preservation and radially symmetric displacement. 

%% file: sections/conclusion.tex
\section{Conclusions}
\label{sec:conclude}

We have presented a patient's head motion-correction mechanism and the constitutive laws that governs the deformation of its constituent soft actuators. The derived model was tested with spheres with size small enough to accommodate a head on a treatment table as shown in \autoref{fig:sofa_setup}. Our results confirm the fidelity of this model given its high accuracy in precise displacements. In future work, we will integrate the soft inverse kinematics relation to the physical build of the system and analyze the properties of head motion correction under the action of a suitable controller. 

%% file: sections/acks.tex
\section{Acknowledgments}
A vote of thanks to  Professor Raymond W. Ogden of The University of Glasgow and Audrey Sedal of The University of Michigan for their feedback on the constitutive model discussed in this work.